\documentclass[10pt, a4paper]{article}
\usepackage{lrec2016}
\usepackage{multibib}
\usepackage{graphicx}

\usepackage{epstopdf}
\usepackage[latin1]{inputenc}

\usepackage{float}
\usepackage{url}
\usepackage{multirow}
\usepackage[table,xcdraw]{xcolor}
\usepackage{tikz}
\usepackage{pgfplots}
\usepackage[normalem]{ulem}
\pgfplotsset{compat=newest}
\usetikzlibrary{decorations.pathmorphing}

\usepackage{tocloft}
\usepackage{titlesec}
\titleformat{\section}
{\large\bfseries\centering}{\thesection.}{1em}{}
 
\renewcommand\thesection{\arabic{section}}
\renewcommand\thesubsection{\thesection.\arabic{subsection}}

\def\checkmark{\tikz\fill[scale=0.4](0,.35) -- (.25,0) -- (0.7,.5) -- (.25,.15) -- cycle;} 

\definecolor{bblue}{HTML}{111111}
\definecolor{rred}{HTML}{CCCCCC}
\definecolor{ggreen}{HTML}{999999}
\definecolor{ppurple}{HTML}{9F4C7C}

\name{St\'{e}phan Tulkens, Lisa Hilte, Elise Lodewyckx, Ben Verhoeven, Walter Daelemans}

\address{CLiPS Research Center, University of Antwerp \\
         Prinsstraat 13, 2000, Antwerpen, Belgium \\
         \{stephan.tulkens, lisa.hilte, ben.verhoeven, walter.daelemans\}@uantwerpen.be, \\elise.lodewyckx@student.uantwerpen.be\\}

\abstract{
We present a dictionary-based approach to racism detection in Dutch social media comments, which were retrieved from two public Belgian social media sites likely to attract racist reactions. These comments were labeled as racist or non-racist by multiple annotators. For our approach, three discourse dictionaries were created: first, we created a dictionary by retrieving possibly racist and more neutral terms from the training data, and then augmenting these with more general words to remove some bias. A second dictionary was created through automatic expansion using a \texttt{word2vec} model trained on a large corpus of general Dutch text. Finally, a third dictionary was created by manually filtering out incorrect expansions. We trained multiple Support Vector Machines, using the distribution of words over the different categories in the dictionaries as features. The best-performing model used the manually cleaned dictionary and obtained an F-score of 0.46 for the racist class on a test set consisting of unseen Dutch comments, retrieved from the same sites used for the training set. The automated expansion of the dictionary only slightly boosted the model's performance, and this increase in performance was not statistically significant. The fact that the coverage of the expanded dictionaries did increase indicates that the words that were automatically added did occur in the corpus, but were not able to meaningfully impact performance. The dictionaries, code, and the procedure for requesting the corpus are available at: \url{https://github.com/clips/hades}.
\newline
\newline
 }

\begin{document}

\title{A Dictionary-based Approach to Racism Detection in Dutch Social Media} 
\author{St\'ephan Tulkens, Lisa Hilte, \\
Elise Lodewyckx, Ben Verhoeven, Walter Daelemans}

\maketitleabstract

\section{Introduction}
Racism is an important issue which is not easily defined, as racist ideas can be expressed in a variety of ways. Furthermore, there is no clear definition of what exactly constitutes a racist utterance; what is racist to one person is highly likely to not be considered racist universally. Additionally, although there exist mechanisms for reporting acts of racism, victims often neglect to do so as they feel that reporting the situation will not solve anything, according to Unia, the Belgian Interfederal Centre for Equal Opportunities.\footnote{\url{http://www.diversiteit.be}} The scope of this issue, however, is currently unknown. Hence, the goal of our system is two-fold: it can be used to shed light on how many racist remarks are not being reported online, and furthermore, the automated detection of racism could provide interesting insights in the linguistic mechanisms used in racist discourse.

In this study, we try to automatically detect racist language in Dutch social media comments, using a dictionary-based approach. We retrieved and annotated comments from two public social media sites which were likely to attract racist reactions according to Unia. We use a Support Vector Machine to automatically classify comments, using handcrafted dictionaries, which were later expanded using automated techniques, as features.

We first discuss previous research on our subject and methodology, and discuss the problem of defining racist language (section \ref{sec:related_research}). Next, we describe our data (section \ref{sec:datasets}). Finally, after discussing the experimental setup (section \ref{sec:experimental_setup}), we present our results (section \ref{sec:results}).

\section{Related Research} \label{sec:related_research}

The classification of racist insults presents us with the problem of giving an adequate definition of racism. More so than in other domains, judging whether an utterance is an act of racism is highly personal and does not easily fit a simple definition. The Belgian anti-racist law forbids discrimination, violence and crime based on physical qualities (like skin color), nationality or ethnicity, but does not mention textual insults based on these qualities.\footnote{\url{http://www.diversiteit.be/de-antiracismewet-van-30-juli-1981}} Hence, this definition is not adequate for our purposes, since it does not include the racist utterances one would find on social media; few utterances that people might perceive as racist are actually punishable by law, as only utterances which explicitly encourage the use of violence are illegal. For this reason, we use a \emph{common sense} definition of racist language, including all negative utterances, negative generalizations and insults concerning ethnicity, nationality, religion and culture. In this, we follow \newcite{paolo2015racist}, \newcite{bonilla2002linguistics} and \newcite{razavi2010offensive}, who show that racism is no longer strictly limited to physical or ethnic qualities, but can also include social and cultural aspects.

Additionally, several authors report linguistic markers of racist discourse; \newcite{vandijk} reports that the number of available topics is greatly restricted when talking about foreigners. \newcite{paolo2015racist}, who performed a qualitative study of posts from Italian social media sites, shows that these chosen topics are typically related to migration, crime and economy. Furthermore, the use of stereotypes and prejudiced statements \cite{reisigl2005discourse,quasthoff}, as well as a heightened occurrence of truth claims \cite{greevy2004classifying,greevy2004text}, are reported as typical characteristics of racist discourse . Finally, racist utterances are said to contain specific words and phrases, i.e. n-grams, significantly more often than neutral texts, like ``our own kind'' and ``white civilization'' \cite{greevy2004classifying,greevy2004text}. 

Stylistically, racist discourse is characterized by a higher rate of certain word classes, like imperatives and adjectives and a higher noun-adjective ratio \cite{paolo2015racist,greevy2004classifying,greevy2004text}. Greevy and Smeaton also report a more frequent use of modals and adverbs, which they link to the higher frequency of truth claims in racist utterances \cite{greevy2004classifying,greevy2004text}. In several studies, pronoun use is reported as an important feature in the detection of racist language. While \newcite{paolo2015racist} reports a high frequency of (especially first person plural) pronouns in racist data, \newcite{vandijk} reports a more general finding: the importance of \emph{us and them} constructions in racist discourse. He explains that they involve a `semantic move with a positive part about Us and a negative part about Them' \cite[p.150]{vandijk}. Using such constructions, one linguistically emphasizes - either deliberately or subconsciously - a divide between groups of people. A strict interpretation implies that even positive utterances about `them' can be perceived as racist, as they can also imply a divide between us and them. In this sense, Van Dijk's definition of racism is subtler, but also broader, than the definition used in our own research: we only count \emph{negative} utterances and generalizations about groups of people as racist.

Our dictionary-based approach is inspired by methods used in previous research, like LIWC (Linguistic Inquiry and Word Count) \cite{pennebaker2001linguistic}. LIWC is a dictionary-based computational tool that counts word frequencies for both grammatical categories (e.g. pronouns) and content-related categories (e.g. negative emotion words). As LIWC uses counts per category instead of individual words' frequencies, it allows for broader generalizations on functionally or semantically related words.

The construction of dictionary categories related to racist discourse (cf. section \ref{subsec:dictionary_construction}) is largely based on linguistic properties of racist language reported in earlier work (see above). Additionally, the categories were adjusted to fit the corpus used in the research, which differs from corpora used in other studies. As our corpus is retrieved from social media sites with an anti-Islamic orientation, we added categories to reflect anti-religious sentiment. The relevant features in this study therefore differ from those reported in other studies, as different words are used to insult different groups of people \cite{greevy2004text}. 

Finally, some other successful quantitative approaches to racism detection that have been used in earlier studies are a bag of words (BoW) approach as well as the analysis of part-of-speech (PoS) tags \cite{greevy2004classifying,greevy2004text}. We leave the addition of these features to future work.

\section{Datasets and Annotations} \label{sec:datasets}

In this section, we describe our data collection, our annotation guidelines (\ref{subsec:annotation_style}) and the results of our annotations (\ref{subsec:annotations_train} and \ref{subsec:annotations_test}).

For our current research we collected a corpus of social media comments, consisting of comments retrieved from Facebook sites which were likely to attract racist reactions in their comments. We specifically targeted two sites: the site of a prominent Belgian anti-Islamic organization, and the site of a Belgian right-wing organization. In both cases the Facebook sites were officially condoned by the organizations, and in the first case  served as a communication platform to organize political gatherings. While both sites, the former more than the latter, explicitly profess to be non-racist, the comments they attracted were still highly critical of foreigners and, predictably, Muslims. This is also the reason we mined \emph{comments} from these sites, and not the posts themselves. While the narrow focus of the sites introduces bias into our data, as the opinions of the people visiting these sites will not reflect the opinions of the general population, they do contain a good proportion of racist to non-racist data. 

\subsection{Annotation Style} \label{subsec:annotation_style}

We annotated the retrieved comments with three different labels: `racist', `non-racist' and `invalid'.

The `racist' label describes comments that contain negative utterances or insults about someone's ethnicity, nationality, religion or culture. This definition also includes utterances which equate, for example, an ethnic group to an extremist group, as well as extreme generalizations. The following examples are comments that were classified as racist:

\begin{enumerate}

\item{Het zijn precies de vreemden die de haat of het racisme opwekken bij de autochtonen.\\ \emph{It is the foreigners that elicit hate and racism from natives.}}

\item{Kan je niets aan doen dat je behoort tot het ras dat nog minder verstand en gevoelens heeft in uw hersenen dan het stinkend gat van een VARKEN ! :-p\\ \emph{You cannot help the fact that you belong to the race that has less intellect and sense in their brains than the smelly behind of a PIG! :-P}}

\item{Wil weer eens lukken dat wij met het vuilste krapuul zitten, ik verschiet er zelfs niet van!\\ \emph{Once again we have to put up with the filthiest scum, it doesn't even surprise me anymore!}}

\end{enumerate}

The label `invalid' was used for comments that were written in languages other than Dutch, or that did not contain any textual information, i.e. comments that solely consist of pictures or links. Before classification, we excluded these from both our training and test set.

The final label, `non-racist', was the default label. If a comment was valid, but could not be considered racist according to our definition, this was the label we used.

\subsection{Training Data} \label{subsec:annotations_train}

To collect the training data, we used \texttt{Pattern}\footnote{\url{http://www.clips.uantwerpen.be/pattern}} \cite{patternref} to scrape the 100 most recent posts from both sites, and then extracted all comments which reacted to these comments. This resulted in 5759 extracted comments: 4880 from the first site and 879 from the second site. The second site attracted a lot less comments on each post, possibly because the site posted more frequently.  In addition to this, the organization behind the first site had been figuring prominently in the news at the time of extraction, which might explain the divide in frequency of comments between the two sites. The corpus was annotated by two annotators, who were both students of comparable age and background. When A and B did not agree on a label, a third annotator, C, was used as a tiebreaker in order to obtain gold-standard labels. Table \ref{gold} shows the gold standard for the training set.

\begin{table}[]
\centering
\begin{tabular}{|l|r|r|}
\hline
           	& \# Train Comments & \# Test Comments \\
\hline\hline
Non-racist  & 4500      & 443        \\
Racist 		& 924       & 164        \\
Invalid    	& 335       & 9          \\
\hline
\end{tabular}

\caption{Gold standard corpus sizes.}
\label{gold}
\end{table}

We calculated inter-annotator agreement using the Kappa score ($\kappa$) \cite{cohen}. On the training corpus, the agreement score was $\kappa$ = 0.60. Annotator A used the racist tag much less often than annotator B. Interestingly, the agreement remains relatively high; 79\% of the comments that A annotated as racist were also annotated as racist by B. Even though B was much more inclined to call utterances racist, A and B still shared a common ground regarding their definition of racism. Examining the comments in detail, we found that the difference can largely be explained by sensitivity to insults and generalizations, as example 4 shows.

\begin{enumerate}
\item[4.]{Oprotten die luizegaards [\emph{sic}] !!! \\ \emph{Throw those lice carriers out!}}
\end{enumerate}

While annotator B considers this utterance to be racist, annotator A does not, as it does not contain a specific reference to an ethnicity, nationality or religion. That is, when not seen in the context of this specific annotation task this sentence would not necessarily be called racist, just insulting.

\subsection{Test data} \label{subsec:annotations_test}

The test corpus was mined in the same way as the training set, at a different point in time. We mined the first 500 and first 116 comments from the first and second site, respectively, which makes the proportion between sites more or less identical to the the proportions in the train corpus. The annotation scheme was identical to the one for the train set, with the difference that C, who previously performed the tiebreak, now became a regular annotator. The first 25\% of each batch of comments, i.e. 125 comments for the first site and 30 comments for the second site, were annotated by all three annotators to compute inter-annotator agreement. The remaining comments were equally divided among annotators. The annotator agreement was $\kappa$ = 0.54 (pairwise average), which is lower than the agreement on the training data. The reason for the lower agreement was that annotator C often did not agree with A and B. Because the pattern of mismatches between the annotators is quite regular, we will now discuss some of the annotations in detail:

\begin{enumerate} 
\item[5.] we kunnen niet iedereen hier binnen laten want dat betekend [\emph{sic}] het einde van de europese beschaving \emph{We cannot let everyone in because that will mean the end of European civilization}

\item[6.] Eigen volk gaat voor, want die vuile manieren van de EU moeten wij vanaf.  Geen EU en geen VN. Waardeloos en tegen onze mensen. (eigen volk.) \\ \emph{Put our own people first, because we need to get rid of the foul manners of the EU. No EU nor UN. Useless and against our people. (own folk.)}

\item[7.] Burgemeester Termont is voor de zwartzakken die kiezen voor hem \\ \emph{Mayor Termont supports the black sacks, as they vote for him}
\end{enumerate}

Annotator C used the `racist' tag more often, which is probably due to the fact that he consistently annotated overt ideological statements related to immigration as `racist', while the other annotators did not. The three examples mentioned above are utterances that C classified as `racist', but A and B classified as `not racist'.

The cause of these consistent differences in annotations might be cultural, as C is from the southern part of the Netherlands, whereas A and B are native to the northern part of Belgium. Some terms are simply misannotated by C because they are Flemish vernacular expressions. For example, \texttt{zwartzak} [black sack], from sentence 7, superficially looks like a derogatory term for a person of color, but actually does not carry this meaning, as it is a slang word for someone who collaborated with the German occupying forces in the Second World War. While this could still be classified as being racist, the point is that C only registered this as a slang word based on skin color, and not a cultural or political term. Finally, it is improbable that the cause of these mismatches is annotator training, as A and B did not discuss their annotations during the task. In addition to this, C functioned as a tiebreaker in the first dataset, and thus already had experience with the nature of the training material.

\section{Experimental Setup} \label{sec:experimental_setup}

In this section, we describe our experimental setup. We will first discuss our dictionary-based approach, describing both the LIWC dictionary we used as well as the construction of dictionaries related to racist discourse (section \ref{subsec:dictionary_construction}). Next, we will describe the preprocessing of the data (section \ref{subsec:preprocessing}).

\subsection{Dictionaries} \label{subsec:dictionary_construction}

\subsubsection{LIWC} \label{subsubsec:LIWC}

In our classification task, we will use the LIWC dictionaries for Dutch\footnote{An exhaustive overview of all categories in the Dutch version of LIWC can be found in \newcite[p. 277-278]{zijlstra2004nederlandse}.} \cite{zijlstra2004nederlandse}. We hypothesize that some of LIWC's word categories can be useful in detecting (implicit) racist discourse, as some of these categories are associated with markers of racist discourse reported in previous research (cf. section \ref{sec:related_research}), including pronouns, negative emotion words, references to others, certainty, religion and curse words.

\subsubsection{Discourse Dictionaries} \label{subsubsec:racism_dicts}

In addition to the Dutch LIWC data, we created a dictionary containing words that specifically relate to racist discourse. We expect a dictionary-based approach in which words are grouped into categories to work well in this case because many of the racist terms used in our corpus were neologisms and hapaxes, like \texttt{halalhoer} (halal prostitute). Alternatively, existing terms are often reused in a ridiculing fashion, e.g. using the word \texttt{mossel} (mussel) to refer to Muslims. The dictionary was created as follows: after annotation, terms pertaining to racist discourse were manually extracted from the training data. These were then grouped into different categories, where most categories have both a neutral and a negative subcategory. The negative subcategory contains explicit insults, while the neutral subcategory contains words that are normally used in a neutral fashion, e.g. \texttt{zwart} (black), \texttt{Marokkaan} (Moroccan), but which might also be used in a more implicit racist discourse; e.g. people that often talk about nationalities or skin color might be participating in a racist us and them discourse. An overview of the categories can be found in Table \ref{tab:cats_racism}.

\begin{table}[]
\centering
\begin{tabular}{|l|l|l|}
\hline
            & Negative   & Neutral    \\
\hline\hline
Skin color  & \checkmark & \checkmark \\
Nationality & \checkmark & \checkmark \\
Religion    & \checkmark & \checkmark \\
Migration   & \checkmark & \checkmark \\
Country     & \checkmark & \checkmark \\
Stereotypes & \checkmark &            \\
Culture     & \checkmark &            \\
Crime       & \checkmark &            \\
Race        & \checkmark &            \\
Disease     & \checkmark &         	  \\
\hline
\end{tabular}
\caption{Overview of the categories in the discourse dictionary}
\label{tab:cats_racism}
\end{table}

After creating the dictionary, we expanded these word lists both manually and automatically. First, we manually added an extensive list of countries, nationalities and languages, to remove some of the bias present in our training corpus. To combat sparsity, and to catch productive compounds which are likely to be used in a racist manner, we added wildcards to the beginning or end of certain words. We used two different wildcards. \texttt{*} is an inclusive wildcard; it matches the word with or without any affixes, e.g. \texttt{moslim*} matches both \texttt{moslim} (Muslim) and \texttt{moslims} (Muslims). \texttt{+} is an exclusive wildcard; it only matches words when an affix is attached, e.g. \texttt{+moslim} will match \texttt{rotmoslim} (Rotten Muslim) but not \texttt{moslim} by itself. In our corpus (which is skewed towards racism), the \texttt{+} will almost always represent a derogatory prefix, which is why it figures more prominently in the negative part of our dictionary. 

A  downside of using dictionaries for the detection of racism, is that they do not include a measure of context. Therefore, a sentence such as ``My brother hated the North African brown rice and lentils we made for dinner''\footnote{We thank an anonymous reviewer for suggesting the sentence.} will be classified as racist,
regardless of the fact that the words above do not occur in a racist context. Approaches based on word unigrams or bigrams face similar problems. This problem is currently partially absolved by the fact that we are working with a corpus skewed towards racism: words like `brown' and `African' are more likely to be racist words in our corpus than in general text.

\subsubsection{Automated Dictionary Expansion}

To broaden the coverage of the categories in our dictionary, we performed dictionary expansion on both the neutral and the negative categories using \texttt{word2vec} \cite{mikolov2013}. \texttt{word2vec} is a collection of models capable of capturing semantic similarity between words based on the sentential contexts in which these words occur. It does so by projecting words into an n-dimensional space, and giving words with similar contexts similar places in this space. Hence, words which are closer to each other as measured by cosine distance, are more similar. Because we observed considerable semantic variation in the insults in our corpus, we expect that dictionary expansion using \texttt{word2vec} will lead to the extraction of previously unknown insults, as we assume that similar insults are used in similar contexts. In parallel, we know that a lot of words belonging to certain semantic categories, such as diseases and animals, can almost invariably be used as insults. 

The expansion proceeded as follows: for each word in the dictionary, we retrieved the five closest words, i.e. the five most similar words, in the n-dimensional space, and added these to the dictionary. Wildcards were not taken into account for this task, e.g. \texttt{*jood} was replaced by \texttt{jood} for the purposes of expansion. As such, the expanded words do not have any wildcards attached to them. For expansion we used the best-performing model from \newcite{tulkens2016}, which is based on a corpus of 3.9 billion words of general Dutch text. Because this \texttt{word2vec} model was trained on general text, the semantic relations contained therein are not based on racist or insulting text, which will improve the coverage of our expanded categories.

After expansion, we manually searched the expanded dictionaries and removed obviously incorrect items. Because the \texttt{word2vec} model also includes some non-Dutch text, e.g. Spanish, some categories were expanded incorrectly. As a result, we have 3 different dictionaries with which we perform our experiments: the original dictionary which was based on the training data, a version which was expanded using \texttt{word2vec}, and a cleaned version of this expanded version. The word frequencies of the dictionaries are given in Table \ref{tab:dict_freq}. An example of expansion is given in Table \ref{example}.

\begin{table}[]
\centering
\begin{tabular}{|l|l|}
\hline
         & \#Words  \\
\hline\hline
Original &  1055       	\\
Expanded &  3845       	\\
Cleaned  &  3532        \\
\hline
\end{tabular}
\caption{Dictionary word frequencies.}
\label{tab:dict_freq}
\end{table}

\begin{table}[]
\centering
\begin{tabular}{|l|l|}
\hline
Dictionary & Example \\
\hline\hline
Original: & mohammed\textbf{*} \\
Expanded: & mohammed*, \textbf{mohamed, mohammad,} \\  
		& \textbf{muhammed, vzmh, hassan} \\
Cleaned: & mohammed*, mohamed, mohammad, \\ 
& muhammed,  \sout{\textbf{vzmh}}, hassan \\
\hline
\end{tabular}
\caption{An example of expansion. The original dictionary only contains a single word. In the expanded version, the \textbf{bold} words have been added. In the third version the words that were \sout{struck through} have been removed.}
\label{example}
\end{table}

\subsection{Preprocessing and Featurization} \label{subsec:preprocessing}

For preprocessing, the text was first tokenized using the Dutch tokenizer from \texttt{Pattern} \cite{patternref}, and then lowercased and split on whitespace, which resulted in lists of words which are appropriate for lexical processing.

Our dictionary-based approach, like LIWC, creates an \emph{n}-dimensional vector of normalized and scaled numbers, where \emph{n} is the number of dictionary categories. These numbers are obtained by dividing the frequency of words in every specific category by the total number of words in the comment.
Because all features are already normalized and scaled, there was no need for further scaling. Furthermore, because the number of features is so small, we did not perform explicit feature selection.

\section{Results and Discussion} \label{sec:results}

\subsection{Performance on the Training Set} \label{subsec:cv_performance}
We estimated the optimal values for the SVM parameters by an exhaustive search through the parameter space, which led to the selection of an RBF kernel with a C value of 1 and a gamma of 0. For the SVM and other experiments, we used the implementation from \texttt{Scikit-Learn} \cite{scikit-learn}. Using cross-validation on the training data, all dictionary-based approaches with lexical categories related to racist discourse significantly outperformed models using only LIWC's general word categories. Since the current research concerns the binary classification of racist utterances, we only report scores for the positive class, i.e. the racist class. When only LIWC-categories were used as features, an F-score of 0.34 (std. dev. 0.07) was obtained for the racist class. When using the original discourse dictionary, we reached an F-score of 0.50 (std. dev. 0.05). Automatic expansion of the categories did not influence performance either (F-score 0.50, std. dev. 0.05). Similar results (0.49 F-score, std. dev. 0.05) were obtained when the expanded racism dictionaries were manually filtered. This result is not surprising, as the original dictionaries were created from the training data, and might form an exhaustive catalog of racist terms in the original corpus. 

Combining the features generated by LIWC with the specific dictionary-based features led to worse results compared to the dictionary-based features by themselves (F-score 0.40, std. dev. 0.07 for the best-performing model). 

Finally, all models based on the dictionary features as well as the combined model outperformed a unigram baseline of 0.36, but the LIWC model did not. We also report a weighted random baseline (WRB), which was outperformed by all models.

\begin{table}[]
\centering
\begin{tabular}{|r|l|l|l|}
\hline
         & P    & R    & F    \\
\hline\hline
Original 	& \textbf{0.42} & 0.61 & \textbf{0.50} \\
Expanded 	& 0.40 & \textbf{0.64} & \textbf{0.50} \\
Cleaned  	& 0.40 & \textbf{0.64} & 0.49 \\
LIWC     	& 0.27 & 0.47  & 0.34 \\
Combined 	& 0.36 & 0.44 & 0.40 \\
Unigram  	& 0.38 & 0.34 & 0.36 \\
\hline
WRB	& 0.27 & 0.27 & 0.27 \\
\hline
\end{tabular}
\caption{Results on the train set. WRB is a weighted random baseline.}
\label{traintable}
\end{table}

\subsection{Testing the Effect of Expansion} \label{subsec:test_runs}

As seen above, the performance of the different models on the train set was comparable, regardless of their expansion. This is due to the creation procedure for the dictionary: because the words in the original dictionary were directly retrieved from the training data, the expanded and cleaned versions might not be able to demonstrate their generalization performance, as most of the racist words from the training data will be included in the original dictionaries as well as the expanded dictionaries. This artifact might disappear in the test set, which was retrieved from the same two sites, but will most likely contain unseen words. These unseen words will not be present in the original dictionary, but could be present in the expanded version.

As Table \ref{testtable} shows, the models obtain largely comparable performance on the test set, and outperform the unigram baseline by a wide margin.

In comparison to previous research, our approach leads to worse results than those of \newcite{greevy2004text}, who report a precision score of 0.93 and a recall score of 0.87, using an SVM with BOW features together with frequency-based term weights. It is, however, difficult to compare these scores to our performance, given that the data, method, and language differ. 

Our best-performing model was based on the expanded and cleaned version of the dictionary, but this model only slightly outperformed the other models. Additionally, we also computed Area Under the Receiving Operator Characteristic Curve (ROC-AUC) scores for all models, also shown in Table \ref{testtable}. ROC-AUC shows the probability of ranking a randomly chosen positive instance above a randomly chosen negative instance, thereby giving an indication of the overall performance of the models. This shows that all dictionaries have comparable AUC scores, and that each dictionary outperforms the unigram baseline. To obtain additional evidence, we computed the statistical significance of performance differences between the models based on the dictionaries and unigram baseline model using approximate randomization testing (ART) \cite{noreen1989computer}.\footnote{We used the implementation by Vincent Van Asch, which is available from the CLiPS website \url{http://www.clips.uantwerpen.be/scripts/art}} An ART test between dictionary models reveals that none of the models had performance differences that were statistically significant. Similarly, all dictionary models outperformed the unigram baseline with statistical significance, with $p$ $<$ 0.01 for the models based on the cleaned and expanded dictionaries, and $p$ $<$ 0.05 for the models based on the original dictionary. 

To get more insight into why the expanded models were not more successful, we calculated dictionary coverage for every dictionary separately on the test set. If the expanded dictionaries do not have increased coverage, the reason for their similar performance is clear: not enough words have been added to affect the performance in any reasonable way. As Table \ref{coverage} indicates, the coverage of the expanded dictionaries did increase, which indicates that the automated expansion, or manual deletion for that matter, contrary to expectations, did not add words that were useful for the classification of racist content. To obtain additional evidence for this claim, we looked at the number of comments that contained  words from the original, cleaned and expanded dictionaries. The coverage in terms of total comments also increased, as well as the absolute number of racist comments that contained the added terms. Because the coverage in number of comments did not increase the performance of the dictionaries, we hypothesize that the terms that were included in the expanded dictionaries were not distributed clearly enough (over racist and neutral texts) to make a difference in the performance on the classification task.

\begin{table}[]
\centering
\begin{tabular}{|r|l|l|l|l|}
\hline
         & P    & R    & F & AUC    \\
\hline\hline
Original & \textbf{0.51} & 0.39 & 0.44 & \textbf{0.63} \\
Expanded & 0.48 & \textbf{0.43} & 0.45 & \textbf{0.63} \\
Cleaned  & 0.49 & \textbf{0.43} & \textbf{0.46} & \textbf{0.63} \\
Unigram  & 0.46 & 0.20 & 0.28 & 0.56 \\
\hline
\end{tabular}
\caption{P, R, F and ROC-AUC scores on the test set.}
\label{testtable}
\end{table}

\begin{table}[]
\centering
\begin{tabular}{|l|r|r|r|r|}
\hline
         & \% Coverage 	& \# comments  & \# racist  \\
\hline\hline
Original &  0.014		&	98		&    43            \\
Expanded &  0.035		&	212		&    82            \\
Cleaned  &  0.034		&	206		&    81            \\
\hline
\end{tabular}
\caption{Coverage of the various dictionaries in vocabulary percentage, number of comments, and number of racist comments.}
\label{coverage}
\end{table}

\section{Conclusions and Future Work} \label{sec:conclusion}

We developed a dictionary-based computational tool for automatic racism detection in Dutch social media comments. These comments were retrieved from public social media sites with an anti-Islamic orientation. The definition of racism we used to annotate the comments therefore includes religious and cultural racism as well, a phenomenon reported on in different studies \cite{paolo2015racist,bonilla2002linguistics,razavi2010offensive}. 

We use a Support Vector Machine to classify comments as racist or not based on the distribution of the comments' words over different word categories related to racist discourse. To evaluate the performance, we used our own annotations as gold standard. The best-performing model obtained an F-score of 0.46 for the racist class on the test set, which is an acceptable decrease in performance compared to cross-validation experiments on the training data (F-score 0.49, std. dev. 0.05). The dictionary used by the model was manually created by retrieving possibly racist and more neutral terms from the training data during annotation. The dictionary was then manually expanded, automatically expanded with a \texttt{word2vec} model and finally manually cleaned, i.e. irrelevant terms that were added automatically were removed. It did not prove useful to use general stylistic or content-based word categories along with the word lists specifically related to racist discourse.

Surprisingly, the expansion of the manually crafted dictionary did not boost the model's performance significantly. In (cross-validated) experiments on the training data, this makes sense, as the words in the different categories are retrieved from the training data itself, artificially making the dictionary very appropriate for the task. In the test runs, however, a better result could be expected from the generalized word lists. The expanded versions of the dictionary had higher overall coverage for the words in the corpus, as well as higher coverage in number of comments and in number of racist comments. This shows that the words that were automatically added, did indeed occur in our corpus. As the model's performance more or less stagnated when using the expanded categories compared to the original ones, we hypothesize that the terms that were automatically added by the \texttt{word2vec} model were irrelevant to the task of discriminating between racist and neutral texts.

In terms of future work, we will expand our research efforts to include more general social media text. Because we currently only use material which was gathered from sites skewed towards racism, the performance of our dictionary might have been artificially heightened, as the words in the dictionary only occur in racist contexts in our corpus. Therefore, including more general social media texts will serve as a good test of the generality of our dictionaries with regards to detecting insulting material. 

\section{Acknowledgments} \label{sec:acknowledgments}
 
We are very grateful towards Leona Erens and Fran\c{c}ois Deleu from Unia for wanting to collaborate with us and for pointing us towards the necessary data. We thank the three anonymous reviewers for their helpful comments and advice.

\section{Supplementary Materials}
The supplementary materials are available at \url{https://github.com/clips/hades}

\section{Bibliographical References}
\bibliographystyle{lrec2016}
\bibliography{references}

\begin{thebibliography}{}

\bibitem[\protect\citename{Bonilla-Silva}2002]{bonilla2002linguistics}
Bonilla-Silva, E.
\newblock (2002).
\newblock The linguistics of color blind racism: How to talk nasty about blacks
  without sounding ``racist".
\newblock {\em Critical Sociology}, 28(1-2):41--64.

\bibitem[\protect\citename{Cohen}1968]{cohen}
Cohen, J.
\newblock (1968).
\newblock Weighted {Kappa}: Nominal scale agreement provision for scaled
  disagreement or partial credit.
\newblock {\em Psychological bulletin}, 70(4):213.

\bibitem[\protect\citename{De~Smedt and Daelemans}2012]{patternref}
De~Smedt, T. and Daelemans, W.
\newblock (2012).
\newblock Pattern for {P}ython.
\newblock {\em The Journal of Machine Learning Research}, 13(1):2063--2067.

\bibitem[\protect\citename{Greevy and Smeaton}2004a]{greevy2004text}
Greevy, E. and Smeaton, S.
\newblock (2004a).
\newblock Text categorization of racist texts using a support vector machine.
\newblock {\em 7 es Journ{\'e}es internationales d'Analyse statistique des
  Donn{\'e}es Textuelles}.

\bibitem[\protect\citename{Greevy and Smeaton}2004b]{greevy2004classifying}
Greevy, E. and Smeaton, A.~F.
\newblock (2004b).
\newblock Classifying racist texts using a support vector machine.
\newblock In {\em Proceedings of the 27th annual international ACM SIGIR
  conference on Research and development in information retrieval}, pages
  468--469. ACM.

\bibitem[\protect\citename{Mikolov \bgroup et al.\egroup }2013]{mikolov2013}
Mikolov, T., Chen, K., Corrado, G., and Dean, J.
\newblock (2013).
\newblock Efficient estimation of word representations in vector space.
\newblock {\em arXiv preprint arXiv:1301.3781}.

\bibitem[\protect\citename{Noreen}1989]{noreen1989computer}
Noreen, E.
\newblock (1989).
\newblock Computer-intensive methods for testing hypotheses: an introduction.

\bibitem[\protect\citename{Orr\`u}2015]{paolo2015racist}
Orr\`u, P.
\newblock (2015).
\newblock Racist discourse on social networks: A discourse analysis of
  {F}acebook posts in {I}taly.
\newblock {\em Rhesis}, 5(1):113--133.

\bibitem[\protect\citename{Pedregosa \bgroup et al.\egroup }2011]{scikit-learn}
Pedregosa, F., Varoquaux, G., Gramfort, A., Michel, V., Thirion, B., Grisel,
  O., Blondel, M., Prettenhofer, P., Weiss, R., Dubourg, V., Vanderplas, J.,
  Passos, A., Cournapeau, D., Brucher, M., Perrot, M., and Duchesnay, E.
\newblock (2011).
\newblock Scikit-learn: Machine learning in {P}ython.
\newblock {\em Journal of Machine Learning Research}, 12:2825--2830.

\bibitem[\protect\citename{Pennebaker \bgroup et al.\egroup
  }2001]{pennebaker2001linguistic}
Pennebaker, J.~W., Francis, M.~E., and Booth, R.~J.
\newblock (2001).
\newblock Linguistic inquiry and word count: {LIWC} 2001.
\newblock {\em Mahway: Lawrence Erlbaum Associates}, 71:2001.

\bibitem[\protect\citename{Quasthoff}1989]{quasthoff}
Quasthoff, U.
\newblock (1989).
\newblock Social prejudice as a resource of power: Towards the functional
  ambivalence of stereotypes.
\newblock {\em Wodak, R.({\'e}d.), Language, Power and Ideology. Amsterdam:
  Benjamins}, pages 181--196.

\bibitem[\protect\citename{Razavi \bgroup et al.\egroup
  }2010]{razavi2010offensive}
Razavi, A.~H., Inkpen, D., Uritsky, S., and Matwin, S.
\newblock (2010).
\newblock Offensive language detection using multi-level classification.
\newblock In {\em Advances in Artificial Intelligence}, pages 16--27. Springer.

\bibitem[\protect\citename{Reisigl and Wodak}2005]{reisigl2005discourse}
Reisigl, M. and Wodak, R.
\newblock (2005).
\newblock {\em Discourse and discrimination: Rhetorics of racism and
  antisemitism}.
\newblock Routledge.

\bibitem[\protect\citename{Tulkens \bgroup et al.\egroup }2016]{tulkens2016}
Tulkens, S., Emmery, C., and Daelemans, W.
\newblock (2016).
\newblock Evaluating unsupervised {Dutch} word embeddings as a linguistic
  resource.
\newblock In {\em Proceedings of the 10th International Conference on Language
  Resources and Evaluation (LREC)}. European Language Resources Association
  (ELRA).

\bibitem[\protect\citename{Van~Dijk}2002]{vandijk}
Van~Dijk, T.~A.
\newblock (2002).
\newblock Discourse and racism.
\newblock {\em The Blackwell companion to racial and ethnic studies}, pages
  145--159.

\bibitem[\protect\citename{Zijlstra \bgroup et al.\egroup
  }2004]{zijlstra2004nederlandse}
Zijlstra, H., Van~Meerveld, T., Van~Middendorp, H., Pennebaker, J.~W., and
  Geenen, R.
\newblock (2004).
\newblock De {N}ederlandse versie van de `linguistic inquiry and word
  count'({LIWC}).
\newblock {\em Gedrag \& gezondheid}, 32:271--281.

\end{thebibliography}

\end{document}